\documentclass[11pt,a4paper]{article}
\usepackage[hyperref]{acl2019}
\usepackage{times}
\usepackage{latexsym}
\usepackage{comment}
\usepackage{amsmath}
\usepackage{amsthm}
\usepackage{amsfonts}
\usepackage{algorithm}
\usepackage{graphicx}
\usepackage{subfigure}
\usepackage[noend]{algpseudocode}
\usepackage{comment}
\usepackage{multirow}
\usepackage{dblfloatfix}
\usepackage{multicol}
\usepackage{url}

\theoremstyle{definition}
\newtheorem{definition}{Definition}[section]

\aclfinalcopy 
\def\aclpaperid{1277} 

\title{Neural Keyphrase Generation via Reinforcement Learning \\with Adaptive Rewards}

\author{Hou Pong Chan\textsuperscript{\rm 1}, 
Wang Chen\textsuperscript{\rm 1}, 
Lu Wang\textsuperscript{\rm 2}, 
{\rm and} Irwin King\textsuperscript{\rm 1} \\
\textsuperscript{\rm 1}The Chinese University of Hong Kong, Shatin, N.T., Hong Kong \\
\textsuperscript{\rm 2}Northeastern Univesity, Boston, MA, USA\\
{\tt \textsuperscript{\rm 1}\{hpchan, wchen, king\}@cse.cuhk.edu.hk}\\
{\tt \textsuperscript{\rm 2}luwang@ccs.neu.edu}
}

\date{}

\begin{document}
\maketitle
\begin{abstract}
Generating keyphrases that summarize the main points of a document is a fundamental task in natural language processing. Although existing generative models are capable of predicting multiple keyphrases for an input document as well as determining the number of keyphrases to generate, they still suffer from the problem of generating too few keyphrases. To address this problem, we propose a reinforcement learning (RL) approach for keyphrase generation, with an adaptive reward function that encourages a model to generate both sufficient and accurate keyphrases. Furthermore, we introduce a new evaluation method that incorporates name variations of the ground-truth keyphrases using the Wikipedia knowledge base. Thus, our evaluation method can more robustly evaluate the quality of predicted keyphrases. Extensive experiments on five real-world datasets of different scales demonstrate that our RL approach consistently and significantly improves the performance of the state-of-the-art generative models with both conventional and new evaluation methods. 
\end{abstract}



\section{Introduction}
\begin{figure}[t]
\centering
\includegraphics[width=0.99\columnwidth]{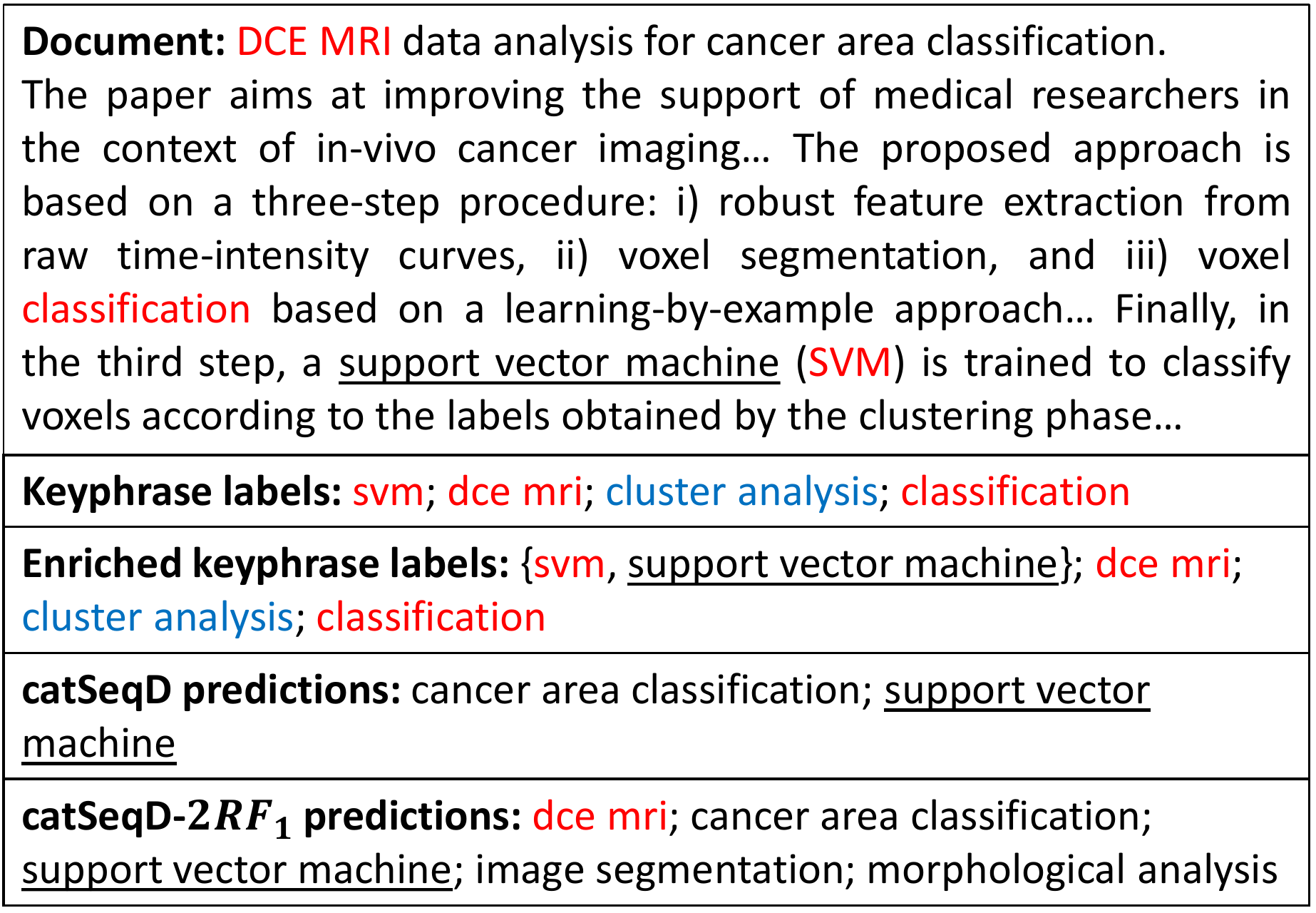}
\caption{
Sample document with keyphrase labels and predicted keyphrases. We use red (blue) color to highlight present (absent) keyphrases. The underlined phrases are name variations of a keyphrase label. ``catSeqD'' is a keyphrase generation model from~\citet{DBLP:journals/corr/diverse-keyphrase}. ``catSeqD-$2RF_{1}$'' denotes the catSeqD model after being trained by our RL approach. The enriched keyphrase labels are based on our new evaluation method.
}
\label{figure:kg-intro-example}
\vspace{-0.15in}
\end{figure}

The task of keyphrase generation aims at predicting a set of keyphrases that convey the core ideas of a document. Figure~\ref{figure:kg-intro-example} shows a sample document and its keyphrase labels. The keyphrases in red color are \textit{present keyphrases} that appear in the document, whereas the blue ones are \textit{absent keyphrases} that do not appear in the input. By distilling the key information of a document into a set of succinct keyphrases, keyphrase generation facilitates a wide variety of downstream applications, including document clustering~\cite{hammouda2005corephrase_doc_clus_app,hulth2006study_doc_clus_app}, opinion mining~\cite{berend2011opinion_opin_app}, and summarization~\cite{zhang2004world_summarize_app, wang2013domain_summarize_app}. 

To produce both present and absent keyphrases, generative methods~\cite{DBLP:conf/acl/MengZHHBC17,DBLP:conf/emnlp/YeW18,DBLP:conf/emnlp/ChenZ0YL18,DBLP:journals/corr/tg-net} are designed to apply the attentional encoder-decoder model~\cite{DBLP:conf/iclr/BahdanauCB14,DBLP:conf/emnlp/LuongPM15} with copy mechanism~\cite{DBLP:conf/acl/GuLLL16,DBLP:conf/acl/SeeLM17} to approach the keyphrase generation task. However, none of the prior models can determine the appropriate number of keyphrases for a document. In reality, the optimal keyphrase count varies, and is dependent on a given document's content. 
To that end, \citet{DBLP:journals/corr/diverse-keyphrase} introduced a training setup in which a generative model can learn to decide the number of keyphrases to predict for a given document and proposed two models. 
Although they provided a more realistic setup, there still exist two drawbacks. 
First, models trained under this setup tend to generate fewer keyphrases than the ground-truth. Our experiments on the largest dataset show that their catSeqD model generates 4.3 keyphrases per document on average, while these documents have 5.3 keyphrase labels on average. Ideally, a model should generate both sufficient and accurate keyphrases. Second, existing evaluation methods rely only on the exact matching of word stems~\cite{DBLP:journals/program/Porter06} to determine whether a predicted phrase matches a ground-truth phrase. For example, given the document in Figure~\ref{figure:kg-intro-example}, if a model generates ``support vector machine'', it will be treated as incorrect since it does not match the word ``svm'' given by the gold-standard labels. 
It is therefore desirable for an evaluation method to consider name variations of a ground-truth keyphrase.  



To address the first limitation, we design an adaptive reward function, $RF_{1}$, that encourages a model to generate both sufficient and accurate keyphrases. Concretely, if the number of generated keyphrases is less than that of the ground-truth, we use recall as the reward, which does not penalize the model for generating incorrect predictions.
If the model generates sufficient keyphrases, we use $F_{1}$ score as the reward, to balance both recall and precision of the predictions. 
To optimize the model towards this non-differentiable reward function, we formulate the task of keyphrase generation as a reinforcement learning (RL) problem and adopt the self-critical policy gradient method~\cite{DBLP:conf/cvpr/RennieMMRG17} as the training procedure. 
Our RL approach is flexible and can be applied to any keyphrase generative model with an encoder-decoder structure. 
In Figure~\ref{figure:kg-intro-example}, we show a prediction result of the catSeqD model~\cite{DBLP:journals/corr/diverse-keyphrase} and another prediction result of the catSeqD model after being trained by our RL approach (catSeqD-$2RF_{1}$). 
This example illustrates that our RL approach encourages the model to generate more correct keyphrases. Perhaps more importantly, the number of generated keyphrases also increases to five, which is closer to the ground-truth number ($5.3$). 




Furthermore, we propose a new evaluation method to tackle the second limitation. For each ground-truth keyphrase, we extract its name variations from various sources. If the word stems of a predicted keyphrase match the word stems of any name variation of a ground-truth keyphrase, it is treated as a correct prediction. For instance, in Figure~\ref{figure:kg-intro-example}, our evaluation method enhances the ``svm'' ground-truth keyphrase with its name variation, ``support vector machine''. Thus, the phrase ``support vector machine'' generated by catSeqD and catSeqD-$2RF_{1}$ will be considered correct, which demonstrates that our evaluation method is more robust than the existing one. 

We conduct extensive experiments to evaluate the performance of our RL approach. Experiment results on five real-world datasets show that our RL approach consistently improves the performance of the state-of-the-art models in terms of $F$-measures. 
Moreover, we analyze the sufficiency of the keyphrases generated by different models. It is observed that models trained by our RL approach generate more absent keyphrases, which is closer to the number of absent ground-truth keyphrases. Finally, we deploy our new evaluation method on the largest keyphrase generation benchmark, and the new evaluation identifies at least one name variation for 14.1\% of the ground-truth keyphrases. 


We summarize our contributions as follows: (1) an RL approach with a novel adaptive reward function that explicitly encourages the model to generate both sufficient and accurate keyphrases; (2) a new evaluation method that considers name variations of the keyphrase labels; and (3) the new state-of-the-art performance on five real-world datasets in a setting where a model is able to determine the number of keyphrases to generate. This is the first work to study RL approach on the keyphrase generation problem.

\section{Related Work}
\subsection{Keyphrase Extraction and Generation}
Traditional extractive methods select important phrases from the document as its keyphrase predictions. Most of them adopt a two-step approach. First, they identify keyphrase candidates from the document by heuristic rules~\cite{DBLP:conf/iconip/WangZH16, DBLP:conf/ausai/LeNS16}. Afterwards, the candidates are either ranked by unsupervised methods~\cite{mihalcea2004textrank, DBLP:conf/aaai/WanX08} or supervised learning algorithms~\cite{medelyan2009human_maui, witten2005kea, DBLP:conf/icadl/NguyenK07}. Other extractive methods apply sequence tagging models~\cite{luan2017scientific_seqlabel, gollapalli2017incorporating_seqlabel, DBLP:conf/emnlp/ZhangWGH16} to identify keyphrases. 
However, extractive methods cannot produce absent keyphrases. 

To predict both present and absent keyphrases for a document, \citet{DBLP:conf/acl/MengZHHBC17} proposed a generative model, CopyRNN, which is composed of an attentional encoder-decoder model~\cite{DBLP:conf/iclr/BahdanauCB14} and a copy mechanism~\cite{DBLP:conf/acl/GuLLL16}.
Lately, multiple extensions to CopyRNN were also presented.
CorrRNN~\cite{DBLP:conf/emnlp/ChenZ0YL18} incorporates the correlation among keyphrases. TG-Net~\cite{DBLP:journals/corr/tg-net} exploits the title information to learn a better representation for an input document. 
\citet{DBLP:journals/corr/abs-1904-03454} leveraged keyphrase extraction models and external knowledge to improve the performance of keyphrase generation. 
\citet{DBLP:conf/emnlp/YeW18} considered a setting where training data is limited, and proposed different semi-supervised methods to enhance the performance. 
All of the above generative models use beam search to over-generate a large number of keyphrases and select the top-$k$ predicted keyphrases as the final predictions, where $k$ is a fixed number. 


Recently, \citet{DBLP:journals/corr/diverse-keyphrase} introduced a setting where a model has to determine the appropriate number of keyphrases for an input document. They proposed a training setup that empowers a generative model to generate variable numbers of keyphrases for different documents. Two new models, catSeq and catSeqD, were described. 
Our work considers the same setting and proposes an RL approach, which is equipped with adaptive rewards to generate sufficient and accurate keyphrases. 
To our best knowledge, this is the first time RL is used for keyphrase generation. 
Besides, we propose a new evaluation method that considers name variations of the keyphrase labels,
a novel contribution to the state-of-the-art.





\subsection{Reinforcement Learning for Text Generation}
Reinforcement learning has been applied to a wide array of text generation tasks, including machine translation~\cite{DBLP:journals/corr/WuSCLNMKCGMKSJL16,DBLP:journals/corr/RanzatoCAZ15}, text summarization~\cite{DBLP:conf/iclr/PaulusXS17,DBLP:conf/ijcai/WangYTZLD18}, and image/video captioning~\cite{DBLP:conf/cvpr/RennieMMRG17,DBLP:conf/iccv/LiuZYG017,DBLP:conf/emnlp/PasunuruB17}. These RL approaches lean on the REINFORCE algorithm~\cite{DBLP:journals/ml/Williams92}, or its variants, to train a generative model towards a non-differentiable reward by minimizing the policy gradient loss.
Different from existing work, our RL approach uses a novel adaptive reward function, which combines the recall and $F_{1}$ score via a hard gate (if-else statement). 


\section{Preliminary}
\subsection{Problem Definition}\label{sec:definition}
We formally define the problem of keyphrase generation as follows. 
Given a document $\mathbf{x}$, output a set of ground-truth keyphrases $\mathcal{Y}=\{\mathbf{y}^{1},\mathbf{y}^{2},\ldots,\mathbf{y}^{|\mathcal{Y}|} \}$. The document $\mathbf{x}$ and each ground-truth keyphrase $\mathbf{y}^{i}$ are sequences of words, i.e., $\mathbf{x}=(x_{1},\ldots,x_{l_{x}})$, and $\mathbf{y}^{i}=(y^{i}_{1},\ldots,y^{i}_{l_{y^i}})$, where $l_{\mathbf{x}}$ and $l_{\mathbf{y}^i}$ denote the numbers of words in $\mathbf{x}$ and $\mathbf{y}^i$ respectively. 
A keyphrase that matches any consecutive subsequence of the document is a present keyphrase, otherwise it is an absent keyphrase. We use $\mathcal{Y}^{p}=\{ y^{p,1},y^{p,2},\ldots,y^{p,|\mathcal{Y}^{p}|} \}$ and $\mathcal{Y}^{a}=\{y^{a,1},y^{a,2},\ldots,y^{a,|\mathcal{Y}^{a}|}\}$ to denote the sets of present and absent ground-truth keyphrases, respectively. 
Thus, the ground-truth keyphrases set can be expressed as $\mathcal{Y}=\mathcal{Y}^{p}\cup \mathcal{Y}^{a}$.

\subsection{Keyphrase Generation Model}\label{sec:model}
In this section, we describe the attentional encoder-decoder model~\cite{DBLP:conf/iclr/BahdanauCB14} 
with copy mechanism~\cite{DBLP:conf/acl/SeeLM17}, which is the backbone of our implementations of the baseline generative models. 

\smallskip
\noindent \textbf{Our training setup.} 
For each document-keyphrases pair $(\mathbf{x},\mathcal{Y})$, we join all the keyphrases in $\mathcal{Y}$ into one output sequence, $\mathbf{y}=\mathbf{y}^{p,1}\wr\mathbf{y}^{p,2}\wr\ldots\wr \mathbf{y}^{p,|\mathcal{Y}^{p}|}\diamond \mathbf{y}^{a,1}\wr\mathbf{y}^{a,2}\wr\ldots\wr \mathbf{y}^{a,|\mathcal{Y}^{a}|}$, where $\diamond$ is a special token that indicates the end of present keyphrases, and $\wr$ is a delimiter between two consecutive present keyphrases or absent keyphrases. Using such $(\mathbf{x},\mathbf{y})$ samples as training data, the encoder-decoder model can learn to generate all the keyphrases in one output sequence and determine the number keyphrases to generate. The only difference with the setup in~\citet{DBLP:journals/corr/diverse-keyphrase} is that we use $\diamond$ to mark the end of present keyphrases, instead of using $\wr$.


\smallskip
\noindent \textbf{Attentional encoder-decoder model.} 
We use a bi-directional Gated-Recurrent Unit (GRU)~\cite{DBLP:conf/emnlp/ChoMGBBSB14} as the encoder. The encoder's $i$-th hidden state is $\mathbf{h}_{i}=[\overrightarrow{\mathbf{h}}_{i};\overleftarrow{\mathbf{h}}_{i}]\in \mathbb{R}^{d_{h}}$. 

A single-layered GRU is adopted as the decoder. At decoding step $t$, 
the decoder hidden state is $\mathbf{s}_{t}=\text{GRU}(\mathbf{e}_{t-1},\mathbf{s}_{t-1})\in \mathbb{R}^{d_{s}}$, where $\mathbf{e_{t-1}}$ is the embedding of the $(t-1)$-th predicted word.
Then we apply the attention layer in~\cite{DBLP:conf/iclr/BahdanauCB14} to compute an attention score $a_{t,i}$ for each of the word $x_{i}$ in the document. The attention scores are next used to compute a context vector $\mathbf{h}^{*}_{t}$ for the document. 
The probability of predicting a word $y_{t}$ from a predefined vocabulary $V$ is defined as 
$P_{V}(y_{t})=\text{softmax}(\mathbf{W}_{V}(\mathbf{W}_{V'}[\mathbf{s}_{t};\mathbf{h}^{*}_{t}]))$.
In this paper, all the $\mathbf{W}$ terms represent trainable parameters and we omit the bias terms for brevity.


\smallskip
\noindent \textbf{Pointer-generator network.} 
To alleviate the out-of-vocabulary (OOV) problem, we adopt the copy mechanism from~\citet{DBLP:conf/acl/SeeLM17}. For each document $\mathbf{x}$, we build a dynamic vocabulary $V_{\mathbf{x}}$ by merging the predefined vocabulary $V$ and all the words that appear in $\mathbf{x}$. Then, the probability of predicting a word $y_{t}$ from the dynamic vocabulary $V_{\mathbf{x}}$ is computed as 
$P_{V_{\mathbf{x}}}(y_{t})=p_{gen}P_{V}(y_{t})+(1-p_{gen})P_{C}(y_{t})$,
where $P_{C}(y_{t})=\sum_{i:x_{i}=y_{t}} a_{t,i}$ is the copy distribution and $p_{gen}=\text{sigmoid}(\mathbf{W}_{g}[\mathbf{h}^{*}_{t};\mathbf{s}_{t};\mathbf{e}_{t-1}])\in[0,1]$ is a soft gate to select between generating a word from the vocabulary $V$ and copying a word from the document.

\smallskip
\noindent \textbf{Maximum likelihood training.}
We use $\theta$ to denote all model parameters and $\mathbf{y}_{1:t-1}$ to denote a sequence $(y_{1},...,y_{t-1})$. Previous work learns the parameters by maximizing the log-likelihood of generating the ground-truth output sequence $\mathbf{y}$, defined as follows,
\begin{align}
\mathcal{L}(\theta)=- \sum_{t=1}^{L_{\mathbf{y}}} \log P_{V_{\mathbf{x}}}(y_{t}|\mathbf{y}_{1:t-1},\mathbf{x};\theta)\text{.}
\end{align}


\section{Reinforcement Learning Formulation}
We formulate the task of keyphrase generation as a reinforcement learning problem, in which an agent interacts with an environment in discrete time steps. At each time step $t=1,\ldots,T$, the agent produces an action (word) $\hat{y}_{t}$ sampled from the policy $\pi(\hat{y}_{t}|\hat{\mathbf{y}}_{1:t-1},\mathbf{x};\theta)$, where $\hat{\mathbf{y}}_{1:t-1}$ denotes the sequence generated by the agent from step $1$ to $t-1$. After that, the environment gives a reward $r_{t}(\hat{\mathbf{y}}_{1:t},\mathcal{Y})$ to the agent and transits to the next step $t+1$ with a new state $\hat{s}_{t+1}=(\hat{\mathbf{y}}_{1:t},\mathbf{x},\mathcal{Y})$. The policy of the agent is a keyphrase generation model, i.e., $\pi(.|\mathbf{\hat{y}}_{1:t-1},\mathbf{x};\theta)=P_{V_{\mathbf{x}}}(.|\mathbf{\hat{y}}_{1:t-1},\mathbf{x};\theta)$.





To improve the sufficiency and accuracy of both present keyphrases and absent keyphrases generated by the agent, we give separate reward signals to present keyphrase predictions and absent keyphrase predictions. Hence, we divide our RL problem into two different stages. In the first stage, we evaluate the agent's performance on extracting present keyphrases. Once the agent generates the `$\diamond$' token, we denote the current time step as $T^{p}$, the environment computes a reward using our adaptive reward function $RF_{1}$ by comparing the generated keyphrases in $\hat{\mathbf{y}}_{1:T^P}$ with the ground-truth present keyphrases  $\mathcal{Y}^{p}$, i.e., $r_{T^{P}}(\hat{\mathbf{y}}_{1:T^{P}},\mathcal{Y})=RF_{1}(\hat{\mathbf{y}}_{1:T^P},\mathcal{Y}^{p})$. Then we enter the second stage, where we evaluate the agent's performance on generating absent keyphrases. Upon generating the EOS token, the environment compares the generated keyphrases in  $\hat{\mathbf{y}}_{T^P+1:T}$ with the ground-truth absent keyphrases $\mathcal{Y}^{a}$ and computes a reward $r_{T}(\hat{\mathbf{y}}_{1:T},\mathcal{Y})=RF_{1}(\hat{\mathbf{y}}_{T^{p}+1:T},\mathcal{Y}^{a})$. After that, the whole process terminates. The reward to the agent is 0 for all other time steps, i.e., $r_{t}(\hat{\mathbf{y}}_{1:t},\mathcal{Y})=0$ for all $t\notin \{T^{p},T\}$. 

Let return $R_{t}(\hat{\mathbf{y}}, \mathcal{Y})$ be the sum of future reward starting from time step $t$, i.e., $R_{t}(\hat{\mathbf{y}},\mathcal{Y}) = \sum_{\tau=t}^{T} r_{\tau}(\hat{\mathbf{y}}_{1:\tau},\mathcal{Y})$, where $\hat{\mathbf{y}}$ denotes the complete sequence generated by the agent, i.e., $\hat{\mathbf{y}}=\hat{\mathbf{y}}_{1:T}$. We then simplify the expression of return into: 
\begin{align} \label{eq:return}
R_{t} = \begin{cases}
RF_{1}(\hat{\mathbf{y}}_{1:T^{P}}, \mathcal{Y}^{p}) + \\
\quad RF_{1}(\hat{\mathbf{y}}_{T^{P}+1:T}, \mathcal{Y}^{a})
& \text{if }1\leq t\leq T^{p}\text{,}\\
RF_{1}(\hat{\mathbf{y}}_{T^{P}+1:T}, \mathcal{Y}^{a})
& \text{if }T^{p} < t\leq T\text{.}
\end{cases}
\end{align}
The goal of the agent is to maximize the expected initial return $\mathbb{E}_{\hat{\mathbf{y}}\sim \pi(.|\mathbf{x};\theta)} R_{1}(\hat{\mathbf{y}}, \mathcal{Y})$, where  $R_{1}(\hat{\mathbf{y}}, \mathcal{Y})=RF_{1}(\hat{\mathbf{y}}_{1:T^{P}}, \mathcal{Y}^{p}) + RF_{1}(\hat{\mathbf{y}}_{T^{P}+1:T}, \mathcal{Y}^{a})$.

\smallskip
\noindent \textbf{Adaptive reward function.} 
To encourage the model to generate sufficient and accurate keyphrases, we define our adaptive reward function $RF_{1}$ as follows. First, let $N$ be the number of predicted keyphrases, and $G$ be the number of ground-truth keyphrases, then
\begin{align} \label{eq:new_f1}
RF_{1} & = \begin{cases}
\text{recall}& \quad \text{if }N<G\text{,}\\
F_{1} & \quad \text{otherwise.}
\end{cases}
\end{align}
If the model generates insufficient number of keyphrases, the reward will be the recall of the predictions. 
Since generating incorrect keyphrases will not decrease the recall, the model is encouraged to produce more keyphrases to boost the reward.
If the model generates a sufficient number of keyphrases, 
the model should be discouraged from over-generating incorrect keyphrases,
thus the $F_{1}$ score is used as the reward, which incorporates the precision of the predicted keyphrases. 


\smallskip
\noindent \textbf{REINFORCE.} To maximize the expected initial return, we define the following loss function:
\begin{align}
L(\theta) = -\mathbb{E}_{\hat{\mathbf{y}}\sim \pi(.|\mathbf{x};\theta)} [R_{1}(\mathbf{\hat{y}}, \mathcal{Y})] \text{.}
\end{align}
According to the REINFORCE learning rule in~\citet{DBLP:journals/ml/Williams92}, the expected gradient of the initial return can be expressed as 
$\nabla_{\theta} L(\theta) = -\mathbb{E}_{\hat{\mathbf{y}}\sim \pi(.|\mathbf{x};\theta)} [\sum_{t=1}^{T} \nabla_{\theta} \log \pi (\hat{y}_{t}|\hat{\mathbf{y}}_{1:t-1},\mathbf{x};\theta) R_{t}]$. 
In practice, we approximate the above expectation using a sample $\hat{\mathbf{y}} \sim \pi(.|\mathbf{x};\theta)$. Moreover, we subtract the return $R_{t}$ by a baseline $B_{t}$, which is a standard technique in RL to reduce the variance of the gradient estimator~\cite{DBLP:books/lib/SuttonB98}. In theory, the baseline can be any function that is independent of the current action $y_{t}$. The gradient $\nabla_{\theta} L$ is then estimated by:
\begin{align}\label{eq:baselined-pg-loss}
\nabla_{\theta} L \approx -\sum_{t=1}^{T} \nabla_{\theta} \log \pi (\hat{y}_{t}|\hat{\mathbf{y}}_{1:t-1}, \mathbf{x}; \theta) (R_{t} - B_{t})\text{.}
\end{align}
Intuitively, the above gradient estimator increases the generation probability of a word $\hat{y}_{t}$ if its return $R_{t}$ is higher than the baseline ($R_{t} - B_{t} > 0$). 


\smallskip
\noindent \textbf{Self-critical sequence training.} 
The main idea of self-critical sequence training~\cite{DBLP:conf/cvpr/RennieMMRG17} is to produce another sequence $\bar{\mathbf{y}}$ from the current model using greedy search algorithm, then use the initial return obtained by $\bar{\mathbf{y}}$ as the baseline. The interpretation is that the gradient estimator increases the probability of a word if it has an advantage over the greedily decoded sequence. 
We apply this idea to our RL problem, which has two different stages. When in the present (absent) keyphrase prediction stage, we want the baseline $B_{t}$ to be the initial return obtained by the greedy sequence $\mathbf{\bar{y}}$ in its present (absent) keyphrase prediction stage. 
Thus, we first let $\bar{T}^{P}$ and $\bar{T}$ be the decoding steps where the greedy search algorithm generates the $\diamond$ token and EOS token, respectively. We then define the baseline\footnote{The value of $B_{t}$ only depends on whether 
`$\diamond$' exists in $\hat{\mathbf{y}}_{1:t-1}$, hence it does not depend on the current action $\hat{y}_{t}$.} as:
\begin{align} \label{eq:baseline}
B_{t} = \begin{cases}
RF_{1}(\bar{\mathbf{y}}_{1:\bar{T}^{P}}, \mathcal{Y}^{p}) + \\
\quad RF_{1}(\bar{\mathbf{y}}_{\bar{T}^{P}+1:\bar{T}}, \mathcal{Y}^{a})
& \text{if }1\leq t\leq T^{p}\text{,}\\
RF_{1}(\bar{\mathbf{y}}_{\bar{T}^{P}+1:\bar{T}}, \mathcal{Y}^{a})
& \text{if } T^{p} < t\leq T\text{.}
\end{cases}
\end{align}
With Eqs.~(\ref{eq:baselined-pg-loss}) and (\ref{eq:baseline}), we can simply perform gradient descent to train a generative model.

\section{New Evaluation Method}
\label{sec:eval_method}
Our new evaluation method maintains a set of name variations $\tilde{\mathbf{y}}^{i}$ for each ground-truth keyphrase $\mathbf{y}^{i}$ of $\mathbf{x}$. If a predicted keyphrase $\hat{\mathbf{y}}^{i}$ matches any name variation of a ground-truth keyphrase, then $\hat{\mathbf{y}}^{i}$ is considered a correct prediction.
A ground-truth keyphrase is also its own name variation.
If there are multiple ground-truth keyphrases in $\mathbf{x}$ that have the same name variations set, we will only keep one of them. 

In our evaluation method, the name variation set of a ground-truth keyphrase may contain both present phrases and absent phrases. In such a case, a ground-truth keyphrase can be matched by a present predicted keyphrase or an absent predicted keyphrase. Thus, this ground-truth keyphrase should be treated as both a present ground-truth keyphrase and an absent ground-truth keyphrase, as shown in the following definition.
\theoremstyle{definition}
\begin{definition}{\textbf{Present (Absent) ground-truth keyphrase.}}
If a name variation set $\tilde{\mathbf{y}}^{i}$ of a ground-truth keyphrase $\mathbf{y}^{i}$ only consists of present (absent) keyphrases, then $\mathbf{y}^{i}$ is a present (absent) ground-truth keyphrase. 
Otherwise, $\mathbf{y}^{i}$ is both a present ground-truth keyphrase and an absent ground-truth keyphrase, i.e., $\mathbf{y}^{i}\in \mathcal{Y}^{p}$ and $\mathbf{y}^{i}\in \mathcal{Y}^{a}$. 
\end{definition}

\subsection{Name Variation Extraction} 
We extract name variations of a ground-truth keyphrase from the following sources: acronyms in the ground-truths, Wikipedia disambiguation pages, and Wikipedia entity titles. The later two sources have also been adopted by entity linking methods~\cite{DBLP:conf/coling/ZhangSTW10, DBLP:conf/ijcai/ZhangSST11} to find name variations. Some examples of extracted name variations are shown in Table~\ref{table:variation-cases}. 

\begin{table}[t]
\centering
\small
\begin{tabular}{l|l}
\hline \hline
\textbf{Ground-truth}         & \textbf{Extracted variations} \\
\hline \hline
pca                           & principal component analysis  \\
ssd                           & solid state drive             \\
op amps                       & operational amplifier         \\
hackday                       & hackathon                     \\
mobile ad hoc networks        & manet                         \\
electronic commerce           & e commerce                    \\
\hline
\end{tabular}
\caption{Examples of name variations extracted by our method for keyphrase labels on the KP20k dataset. }
\label{table:variation-cases}
\vspace{-0.15in}
\end{table}

\smallskip
\noindent \textbf{Acronyms in the ground-truths.} 
We found that some of the ground-truth keyphrases have included an acronym at the end of the string, e.g.,``principal component analysis (pca)''. Thus, we adopt the following simple rule to extract an acronym from a ground-truth keyphrase. If a ground-truth keyphrase ends with a pair of parentheses, we will extract the phrase inside the pair, e.g., ``pca'', as one of the name variations.

\smallskip
\noindent \textbf{Wikipedia entity titles.} 
An entity page in Wikipedia provides the information of an entity, and the page title represents an unambiguous name variation of that entity.
For example, a search for ``solid state disk'' on Wikipedia will be redirected to the entity page of ``solid state drive''. In such case, the title ``solid state drive'' is a name variation of ``solid state disk''.

\smallskip
\noindent \textbf{Wikipedia disambiguation pages.} 
A disambiguation page helps users find the correct entity page when the input query refers to more than one entity in Wikipedia. It contains a list of entity pages that the query refers to. 
For example, a keyphrase of ``ssd'' may refer to the entity ``solid state drive'' or ``sterol-sensing domain'' in Wikipedia. To find the correct entity page for a keyphrase, we iterate through this list of possible entities. If an entity title is present in a document, we assume it is the entity that the keyphrase refers to. For example, if a document $\mathbf{x}$ contains ``solid state drive'', we will assume that the keyphrase ``ssd'' refers to this entity. 

\section{Experiments}
We first report the performance of different models using the conventional evaluation method. Afterwards, we present the results based on our new evaluation method. All experiments are repeated for three times using different random seeds and the averaged results are reported. 
The source code and the enriched evaluation set are released to the public\footnote{Source code and evaluation set are available at https://github.com/kenchan0226/keyphrase-generation-rl}. Sample output is shown in Figure~\ref{figure:kg-intro-example}. 

\subsection{Datasets}
We conduct experiments on five scientific article datasets, including \textbf{KP20k}~\cite{DBLP:conf/acl/MengZHHBC17}, \textbf{Inspec}~\cite{Hulth2003inspec}, \textbf{Krapivin}~\cite{Krapivin2009LargeDF}, \textbf{NUS}~\cite{Nguyen2007NUS}, and \textbf{SemEval}~\cite{Kim2010SemEval}.
Each sample from these datasets consists of the title, abstract, and keyphrases of a scientific article. We concatenate the title and abstract as an input document, and use the assigned keyphrases as keyphrase labels. 
Following the setup in ~\cite{DBLP:conf/acl/MengZHHBC17, DBLP:journals/corr/diverse-keyphrase, DBLP:journals/corr/tg-net}, we use the training set of the largest dataset, KP20k, for model training and the testing sets of all five datasets to evaluate the performance of a generative model. From the training set of KP20k, we remove all articles that are duplicated in itself, either in the KP20k validation set, or in any of the five testing sets. After the cleanup, the KP20k dataset contains 509,818 training samples, 20,000 validation samples, and 20,000 testing samples.

\subsection{Evaluation Metrics}
The performance of a model is typically evaluated by comparing the top $k$ predicted keyphrases with the ground-truth keyphrases.
The evaluation cutoff $k$ can be either a fixed number or a variable. 
Most previous work~\cite{DBLP:conf/acl/MengZHHBC17,DBLP:conf/emnlp/YeW18,DBLP:conf/emnlp/ChenZ0YL18,DBLP:journals/corr/tg-net} adopted evaluation metrics with fixed evaluation cutoffs, e.g., $F_{1}@5$. 
Recently, \citet{DBLP:journals/corr/diverse-keyphrase} proposed a new evaluation metric, $F_{1}@M$, which has a variable evaluation cutoff. 
$F_{1}@M$ compares all the keyphrases predicted by the model with the ground-truth to compute an $F_{1}$ score, i.e., $k = $ number of predictions. 
It can also be interpreted as the original $F_{1}$ score with no evaluation cutoff. 


We evaluate the performance of a model using a metric with a variable cutoff and a metric with a fixed cutoff, namely, $F_{1}@M$ and $F_{1}@5$. Marco average is deployed to aggregate the evaluation scores for all testing samples. We apply Porter Stemmer before determining whether two phrases are matched. Our implementation of $F_{1}@5$ is different from that of~\citet{DBLP:journals/corr/diverse-keyphrase}. Specifically, when computing $F_{1}@5$, if a model generates less than five predictions, we append random wrong answers to the prediction until it reaches five predictions\footnote{The implementation in ~\citet{DBLP:journals/corr/diverse-keyphrase} sets $F_{1}@5=F_{1}@M$ for such samples.}. 
The rationale is to avoid producing similar $F_{1}@5$ and $F_{1}@M$, when a model (e.g., catSeq) generates less than five keyphrases,
as shown in the Table 2 of~\citet{DBLP:journals/corr/diverse-keyphrase}. 


\begin{table*}[t]
\centering
\small
\begin{tabular}{ l | c c | c c | c c | c c | c c}
\hline \hline
\multicolumn{1}{c|}{\multirow{2}{*}{\textbf{Model}}} & \multicolumn{2}{c|}{\textbf{Inspec}} & \multicolumn{2}{c|}{\textbf{Krapivin}} & \multicolumn{2}{c|}{\textbf{NUS}} & \multicolumn{2}{c|}{\textbf{SemEval}} & \multicolumn{2}{c}{\textbf{KP20k}} \\
\multicolumn{1}{c|}{} & $F_{1}@M$   & $F_{1}@5$   & $F_{1}@M$    & $F_{1}@5$    & $F_{1}@M$  & $F_{1}@5$ & $F_{1}@M$    & $F_{1}@5$   & $F_{1}@M$   & $F_{1}@5$  \\
\hline \hline
catSeq      &0.262 & 0.225 & 0.354 & 0.269 & 0.397 & 0.323& 0.283 & 0.242 & 0.367 & 0.291 \\
catSeqD     &0.263 & 0.219 & 0.349 & 0.264 & 0.394 & 0.321& 0.274 & 0.233 & 0.363 & 0.285 \\
catSeqCorr   &0.269 & 0.227 & 0.349 & 0.265 & 0.390 & 0.319& 0.290 & 0.246 & 0.365 & 0.289 \\
catSeqTG    &0.270 & 0.229 & 0.366 & 0.282 & 0.393 & 0.325& 0.290 & 0.246 & 0.366 & 0.292 \\
\hline
catSeq-$2RF_{1}$     &0.300 &0.250 & 0.362 & 0.287 &0.426 & 0.364& 0.327 & 0.285& 0.383& 0.310 \\
catSeqD-$2RF_{1}$    &0.292 &0.242 & 0.360 & 0.282 &0.419 & 0.353& 0.316 & 0.272& 0.379& 0.305 \\
catSeqCorr-$2RF_{1}$  &0.291 &0.240 & \textbf{0.369} & 0.286 &0.414 & 0.349& 0.322 & 0.278& 0.382& 0.308 \\
catSeqTG-$2RF_{1}$   &\textbf{0.301} &\textbf{0.253} & \textbf{0.369} & \textbf{0.300} &\textbf{0.433} & \textbf{0.375}& \textbf{0.329} & \textbf{0.287}& \textbf{0.386}& \textbf{0.321} \\
\hline
\end{tabular}
\caption{
Results of present keyphrase prediction on five datasets. Suffix ``-$2RF_{1}$'' denotes that a model is trained by our reinforcement learning approach.}
\label{table:present-result}
\end{table*}

\subsection{Baseline and Deep Reinforced Models}

We train four baseline generative models using maximum-likelihood loss. These models include \textbf{catSeq}, \textbf{catSeqD}~\cite{DBLP:journals/corr/diverse-keyphrase}, \mbox{\textbf{catSeqCorr}}~\cite{DBLP:conf/emnlp/ChenZ0YL18}, and \mbox{\textbf{catSeqTG}}~\cite{DBLP:journals/corr/tg-net}. 
For all baselines, we use the method in~\citet{DBLP:journals/corr/diverse-keyphrase} to prepare the training data, by concatenating all keyphrases into one output sequence.
With this setup, all baselines can determine the number of keyphrases to generate. 
The catSeqCorr and catSeqTG models are the CorrRNN~\cite{DBLP:conf/emnlp/ChenZ0YL18} and TG-Net~\cite{DBLP:journals/corr/tg-net} models trained under this setup, respectively. 

For the reinforced models, we follow the method in Section~\ref{sec:model} to concatenate keyphrases. We first pre-train each baseline model using maximum-likelihood loss, and then apply our RL approach to train each of them. We use a suffix ``-$2RF_{1}$'' to indicate that a generative model is fine-tuned by our RL algorithm, e.g., catSeq-$2RF_{1}$. 



\subsection{Implementation Details}
Following~\cite{DBLP:journals/corr/diverse-keyphrase}, we use greedy search (beam search with beam width 1) as the decoding algorithm during testing. We do not apply the Porter Stemmer to the keyphrase labels in the SemEval testing dataset because they have already been stemmed. We remove all the duplicated keyphrases from the predictions before computing an evaluation score.
The following steps are applied to preprocess all the datasets. We lowercase all characters, replace all the digits with a special token $\langle digit \rangle$, and perform tokenization. Following~\cite{DBLP:journals/corr/diverse-keyphrase}, for each document, we sort all the present keyphrase labels according to their order of the first occurrence in the document. The absent keyphrase labels are then appended at the end of present keyphrase labels. We do not rearrange the order among the absent keyphrases.

The vocabulary $V$ is defined as the most frequent 50,002 words, i.e., $|V|=50002$. We train all the word embeddings from scratch with a hidden size of 100.
The hidden size of encoder $d_{h}$ and the hidden size of decoder $d_{s}$ are both set to 300. The followings are the dimensions of the $\mathbf{W}$ terms: $\mathbf{W}_{V}\in \mathbb{R}^{|V|\times d_{s}}$, $\mathbf{W}_{V'}\in \mathbb{R}^{d_{s}\times (d_{h}+d_{s})}$, $\mathbf{W}_{g}\in \mathbb{R}^{1\times(d_{h}+d_{s}+100)}$.
The encoder bi-GRU has only one layer. The initial state of the decoder GRU is set to $[\overrightarrow{\mathbf{h}}_{L_{\mathbf{x}}}; \overleftarrow{\mathbf{h}}_1]$. For all other model parameters of the baseline models, we follow the dimensions specified by their corresponding papers~\cite{DBLP:journals/corr/diverse-keyphrase, DBLP:conf/emnlp/ChenZ0YL18, DBLP:journals/corr/tg-net}. We initialize all the model parameters using a uniform distribution within the interval $[-0.1,0.1]$. During training, we use a dropout rate of 0.1 and gradient clipping of 1.0.

For maximum-likelihood training (as well as pretraining), we use the Adam optimization algorithm~\cite{Kingma2014Adam} with a batch size of 12 and an initial learning rate of 0.001. We evaluate the validation perplexity of a model for every 4000 iterations. We reduce the learning rate by half if the validation perplexity (ppl) stops dropping for one check-point and stop the training when the validation ppl stops dropping for three contiguous check-points. We also use teaching-forcing during the training.

For RL training, we use the Adam optimization algorithm~\cite{Kingma2014Adam} with a batch size of 32 and an initial learning rate of 0.00005. We evaluate the validation initial return of a model for every 4000 iterations. We stop the training when the validation initial return stops increasing for three contiguous check-points. If the model generates more than one `$\diamond$' segmenter, we will only keep the first one and remove the duplicates. If the model does not generate the `$\diamond$' segmenter, we will manually insert a `$\diamond$' segmenter to the first position of the generated sequence. 

\begin{table*}[t]
\centering
\small
\begin{tabular}{l | l l | l l | l l | l l | l l}
\hline \hline
\multicolumn{1}{c|}{\multirow{2}{*}{\textbf{Model}}} & \multicolumn{2}{c|}{\textbf{Inspec}} & \multicolumn{2}{c|}{\textbf{Krapivin}} & \multicolumn{2}{c|}{\textbf{NUS}} & \multicolumn{2}{c|}{\textbf{SemEval}} & \multicolumn{2}{c}{\textbf{KP20k}} \\
\multicolumn{1}{c|}{} & $F_{1}@M$   & $F_{1}@5$   & $F_{1}@M$    & $F_{1}@5$    & $F_{1}@M$  & $F_{1}@5$ & $F_{1}@M$    & $F_{1}@5$   & $F_{1}@M$   & $F_{1}@5$  \\
\hline \hline
catSeq      &0.008 & 0.004 & 0.036 & 0.018 & 0.028 & 0.016& 0.028 & 0.020 & 0.032 & 0.015 \\
catSeqD     &0.011 & 0.007 & 0.037 & 0.018 & 0.024 & 0.014& 0.024 & 0.016 & 0.031 & 0.015 \\
catSeqCorr   &0.009 & 0.005 & 0.038 & 0.020 & 0.024 & 0.014& 0.026 & 0.018 & 0.032 & 0.015 \\
catSeqTG    &0.011 & 0.005 & 0.034 & 0.018 & 0.018 & 0.011& 0.027 & 0.019 & 0.032 & 0.015 \\
\hline
catSeq-$2RF_{1}$     &0.017 &0.009 & 0.046 & 0.026 &0.031 & 0.019& 0.027 & 0.018& 0.047& 0.024 \\
catSeqD-$2RF_{1}$    &\textbf{0.021} &0.010 & 0.048 & 0.026 &\textbf{0.037} & \textbf{0.022}& 0.030 & \textbf{0.021}& 0.046& 0.023 \\
catSeqCorr-$2RF_{1}$  &0.020 &0.010 & 0.040 & 0.022 &\textbf{0.037} & \textbf{0.022}& \textbf{0.031} & \textbf{0.021}& 0.045& 0.022 \\
catSeqTG-$2RF_{1}$   &\textbf{0.021} &\textbf{0.012} & \textbf{0.053} & \textbf{0.030} &0.031 & 0.019& 0.030 & \textbf{0.021}& \textbf{0.050}& \textbf{0.027} \\
\hline
\end{tabular}
\caption{Results of absent keyphrase prediction on five datasets.}
\label{table:absent-result}
\vspace{-0.1in}
\end{table*}

\begin{table}[t]
\centering
\small
\begin{tabular}{l|cc|cc}
\hline \hline
\multicolumn{1}{c|}{\multirow{2}{*}{\textbf{Model}}} & \multicolumn{2}{c|}{\textbf{Present}} & \multicolumn{2}{c}{\textbf{Absent}} \\
\multicolumn{1}{c|}{}                                & MAE         & Avg. \#        & MAE           & Avg. \#        \\
\hline \hline
oracle                                               & 0.000     &  2.837       &  0.000        &  2.432         \\
\hline
catSeq                                               & 2.271     &  3.781       &  1.943        &  0.659         \\
catSeqD                                              & 2.225     &  3.694       &  1.961        &  0.629         \\
catSeqCorr                                            & 2.292     &  3.790       &  1.914        &  0.703          \\
catSeqTG                                             & 2.276     &  3.780       &  1.956        &  0.638          \\
\hline
catSeq-$2RF_{1}$                                               & 2.118     &  3.733       &  1.494        &  1.574         \\
catSeqD-$2RF_{1}$                                              & \textbf{2.087}     &  \textbf{3.666}       &  1.541        &  1.455         \\
catSeqCorr-$2RF_{1}$                                            & 2.107     &  3.696       &  1.557        &  1.409          \\
catSeqTG-$2RF_{1}$                                         & 2.204     &  3.865       &  \textbf{1.439}        &  \textbf{1.749}          \\
\hline
\end{tabular}
\caption{The abilities of predicting the correct number of keyphrases on the KP20k dataset. MAE denotes the mean absolute error (the lower the better), Avg. \# denotes the average number of generated keyphrases per document.}
\label{table:number-of-keyphrases}
\vspace{-0.15in}
\end{table}

\subsection{Main Results}
In this section, we evaluate the performance of present keyphrase prediction and absent keyphrase prediction separately. 
The evaluation results of different models on predicting present keyphrases are shown in Table~\ref{table:present-result}. We observe that our reinforcement learning algorithm consistently improves the keyphrase extraction ability of all baseline generative models by a large margin. On the largest dataset KP20k, all reinforced models obtain significantly higher $F_{1}@5$ and $F_{1}@M$ ($p<0.02$, $t$-test) than the baseline models. 

We then evaluate the performance of different models on predicting absent keyphrases. Table~\ref{table:absent-result} suggests that our RL algorithm enhances the performance of all baseline generative models on most datasets, and maintains the performance of baseline methods on the SemEval dataset. Note that predicting absent keyphrases for a document is an extremely challenging task~\cite{DBLP:journals/corr/diverse-keyphrase}, thus the significantly lower scores than those of present keyphrase prediction.

\subsection{Number of Generated Keyphrases}
We analyze the abilities of different models to predict the appropriate number of keyphrases. All duplicated keyphrases are removed during preprocessing. 
We first measure the mean absolute error (MAE) between the number of generated keyphrases and the number of ground-truth keyphrases for all documents in the KP20k dataset. We also report the average number of generated keyphrases per document, denoted as ``Avg. \#''.  
The results are shown in Table~\ref{table:number-of-keyphrases}, where oracle is a model that always generates the ground-truth keyphrases. 
The resultant MAEs demonstrate that our deep reinforced models notably outperform the baselines on predicting the number of absent keyphrases and slightly outperform the baselines on predicting the number of present keyphrases. 
Moreover, our deep reinforced models generate significantly more absent keyphrases than the baselines ($p<0.02$, $t$-test). The main reason is that the baseline models can only generate very few absent keyphrases, whereas our RL approach uses recall as the reward and encourages the model to generate more absent keyphrases. 
Besides, the baseline models and our reinforced models generate similar numbers of present keyphrases, while our reinforced models achieve notably higher $F$-measures, implying that our methods generate present keyphrases more accurately than the baselines. 


\begin{table}[t]
\small
\centering
\begin{tabular}{l|ll|ll}
\hline \hline
\multicolumn{1}{c|}{\multirow{2}{*}{\textbf{Model}}} & \multicolumn{2}{c|}{\textbf{Present}} & \multicolumn{2}{c}{\textbf{Absent}} \\
\multicolumn{1}{c|}{}                                & $F_{1}@M$         & $F_{1}@5$         & $F_{1}@M$         & $F_{1}@5$        \\
\hline \hline
catSeq                                               & 0.367     &  0.291       &  0.032        &  0.015         \\
catSeq-$RF_{1}$                                           & 0.380     &  0.336       &  0.006        &  0.003         \\
catSeq-$2F_{1}$                                           & 0.378     &  0.278       &  0.042        &  0.020         \\
catSeq-$2RF_{1}$                                          & 0.383     &  0.310       &  0.047        &  0.024         \\
\hline
\end{tabular}
\caption{Ablation study on the KP20k dataset. 
Suffix ``-$2RF_{1}$'' denotes our full RL approach. Suffix ``-$2F_{1}$'' denotes that we replace our adaptive $RF_{1}$ reward function in the full approach by an $F_{1}$ reward function. Suffix ``-$RF_{1}$'' denotes that we replace the two separate $RF_{1}$ reward signals in our full approach with only one $RF_{1}$ reward signal for all the generated keyphrases.}
\label{table:ablation}
\vspace{-0.1in}
\end{table}

\subsection{Ablation Study}
We conduct an ablation study to further analyze our reinforcement learning algorithm. The results are reported in Table~\ref{table:ablation}. 

\begin{table}[t]
\centering
\small
\resizebox{\columnwidth}{!}{
\begin{tabular}{l|ll|ll}
\hline \hline
\multicolumn{1}{c|}{\multirow{3}{*}{\textbf{Model}}} & \multicolumn{2}{c|}{\textbf{Present}}                         & \multicolumn{2}{c}{\textbf{Absent}}                         \\
\multicolumn{1}{c|}{}                                & $F_{1}@M$                    & \multicolumn{1}{l|}{$F_{1}@M$} & $F_{1}@M$                    & $F_{1}@M$                     \\
\multicolumn{1}{c|}{}                                & \multicolumn{1}{c}{old} & \multicolumn{1}{c|}{new}  & \multicolumn{1}{c}{old} & \multicolumn{1}{c}{new} \\
\hline \hline
catSeq                                               & 0.367     &  0.376       &  0.032        &  0.034         \\
catSeqD                                          & 0.363     &  0.372       &  0.031        &  0.033         \\
catSeqCorr                                               & 0.365     &  0.375       &  0.032        &  0.034         \\
catSeqTG                                          & 0.366     &  0.374       &  0.032        &  0.033         \\
\hline
catSeq-$2RF_{1}$                                               & 0.383     &  0.396       &  0.047        &  0.054         \\
catSeqD-$2RF_{1}$                                          & 0.379     &  0.390       &  0.046        &  0.052         \\
catSeqCorr-$2RF_{1}$                                               & 0.382     &  0.393       &  0.045        &  0.051         \\
catSeqTG-$2RF_{1}$                                          & 0.386     &  0.398       &  0.050        &  0.056         \\
\hline
\end{tabular}
}
\caption{
Keyphrase prediction results on the KP20k dataset with our new evaluation method.}
\label{table:new-evaluation}
\vspace{-0.1in}
\end{table}

\smallskip
\noindent \textbf{Single Reward vs. Separate Rewards}. 
To verify the effectiveness of separately rewarding present and absent keyphrases, we train the catSeq model using another RL algorithm which only gives one reward for all generated keyphrases without distinguishing present keyphrases and absent keyphrases. We use ``catSeq-$RF_{1}$'' to denote such a method. As seen in Table~\ref{table:ablation}, although the performance of catSeq-$RF_{1}$ is competitive to catSeq-$2RF_{1}$ on predicting present keyphrases, it yields an extremely poor performance on absent keyphrase prediction. 
We analyze the cause as follows. During the training process of catSeq-$RF_{1}$, generating a correct present keyphrase or a correct absent keyphrase leads to the same degree of improvement in the return at every time step. Since producing a correct present keyphrase is an easier task, the model tends to generate present keyphrases only. 


\smallskip
\noindent \textbf{Alternative reward function.} 
We implement a variant of our RL algorithm by replacing the adaptive $RF_{1}$ reward function with an $F_{1}$ score function (indicated with a suffix ``-$2F_{1}$'' in the result table). By comparing the last two rows in Table~\ref{table:ablation}, we observe that our $RF_{1}$ reward function slightly outperforms the $F_{1}$ reward function. 

\subsection{Analysis of New Evaluation Method}
We extract name variations for all keyphrase labels in the testing set of KP20k dataset, following the methodology in Section~\ref{sec:eval_method}. 
Our method extracts at least one additional name variation for 14.1\% of the ground-truth keyphrases. For these enhanced keyphrases, the average number of name variations extracted is 1.01. 
Among all extracted name variations, 14.1\% come from the acronym in the ground-truth, 28.2\% from the Wikipedia disambiguation pages, and the remaining 61.6\% from Wikipedia entity page titles. 

We use our new evaluation method to evaluate the performance of different keyphrase generation models, and compare with the existing evaluation method. Table~\ref{table:new-evaluation} shows that for all generative models, the evaluation scores computed by our method are higher than those computed by prior method. This demonstrates that our proposed evaluation successfully captures name variations of ground-truth keyphrases generated by different models, and can therefore evaluate the quality of generated keyphrases in a more robust manner.

\section{Conclusion and Future Work}
In this work, we propose the first RL approach to the task of keyphrase generation. In our RL approach, we introduce an adaptive reward function $RF_{1}$, which encourages the model to generate both sufficient and accurate keyphrases. Empirical studies on real data demonstrate that our deep reinforced models consistently outperform the current state-of-the-art models. 
In addition, we propose a novel evaluation method which incorporates name variations of the ground-truth keyphrases. As a result, it can more robustly evaluate the quality of generated keyphrases. One potential future direction is to investigate the performance of other encoder-decoder architectures on keyphrase generation such as  Transformer~\cite{DBLP:conf/nips/VaswaniSPUJGKP17} with multi-head attention module~\cite{DBLP:conf/emnlp/LiTYLZ18,DBLP:conf/uai/ZhangSXMKY18}. Another interesting direction is to apply our RL approach on the microblog hashtag annotation problem~\cite{wang2019microblog,DBLP:conf/ijcai/GongZ16,DBLP:conf/naacl/ZhangLSZ18}. 




\section*{Acknowledgments}
The work described in this paper was partially supported by the Research Grants Council of the Hong Kong Special Administrative Region, China (No.~CUHK~14208815 of the General Research Fund) and Meitu~(No.~7010445). 
Lu Wang is supported in part by National Science Foundation through Grants IIS-1566382 and IIS-1813341, and by the Office of the Director of National Intelligence (ODNI), Intelligence Advanced Research Projects Activity (IARPA), via contract \# FA8650-17-C-9116. The views and conclusions contained herein are those of the authors and should not be interpreted as necessarily representing the official policies, either expressed or implied, of ODNI, IARPA, or the U.S. Government. The U.S. Government is authorized to reproduce and distribute reprints for governmental purposes notwithstanding any copyright annotation therein. 
We would like to thank Jiani Zhang, and the three anonymous reviewers for their comments.

\bibliography{acl2019}

\begin{thebibliography}{49}
\expandafter\ifx\csname natexlab\endcsname\relax\def\natexlab#1{#1}\fi

\bibitem[{Bahdanau et~al.(2014)Bahdanau, Cho, and
  Bengio}]{DBLP:conf/iclr/BahdanauCB14}
Dzmitry Bahdanau, Kyunghyun Cho, and Yoshua Bengio. 2014.
\newblock Neural machine translation by jointly learning to align and
  translate.
\newblock In \emph{International Conference on Learning Representations
  ({ICLR})}.

\bibitem[{Berend(2011)}]{berend2011opinion_opin_app}
G{\'{a}}bor Berend. 2011.
\newblock \href {http://aclweb.org/anthology/I/I11/I11-1130.pdf} {Opinion
  expression mining by exploiting keyphrase extraction}.
\newblock In \emph{Fifth International Joint Conference on Natural Language
  Processing, {IJCNLP} 2011, Chiang Mai, Thailand, November 8-13, 2011}, pages
  1162--1170.

\bibitem[{Chen et~al.(2018{\natexlab{a}})Chen, Zhang, Wu, Yan, and
  Li}]{DBLP:conf/emnlp/ChenZ0YL18}
Jun Chen, Xiaoming Zhang, Yu~Wu, Zhao Yan, and Zhoujun Li. 2018{\natexlab{a}}.
\newblock \href {https://aclanthology.info/papers/D18-1439/d18-1439} {Keyphrase
  generation with correlation constraints}.
\newblock In \emph{Proceedings of the 2018 Conference on Empirical Methods in
  Natural Language Processing, Brussels, Belgium, October 31 - November 4,
  2018}, pages 4057--4066.

\bibitem[{Chen et~al.(2019)Chen, Chan, Li, Bing, and
  King}]{DBLP:journals/corr/abs-1904-03454}
Wang Chen, Hou~Pong Chan, Piji Li, Lidong Bing, and Irwin King. 2019.
\newblock \href {http://arxiv.org/abs/1904.03454} {An integrated approach for
  keyphrase generation via exploring the power of retrieval and extraction}.
\newblock \emph{CoRR}, abs/1904.03454.

\bibitem[{Chen et~al.(2018{\natexlab{b}})Chen, Gao, Zhang, King, and
  Lyu}]{DBLP:journals/corr/tg-net}
Wang Chen, Yifan Gao, Jiani Zhang, Irwin King, and Michael~R. Lyu.
  2018{\natexlab{b}}.
\newblock \href {http://arxiv.org/abs/1808.08575} {Title-guided encoding for
  keyphrase generation}.
\newblock \emph{CoRR}, abs/1808.08575.

\bibitem[{Cho et~al.(2014)Cho, van Merrienboer, G{\"{u}}l{\c{c}}ehre, Bahdanau,
  Bougares, Schwenk, and Bengio}]{DBLP:conf/emnlp/ChoMGBBSB14}
Kyunghyun Cho, Bart van Merrienboer, {\c{C}}aglar G{\"{u}}l{\c{c}}ehre, Dzmitry
  Bahdanau, Fethi Bougares, Holger Schwenk, and Yoshua Bengio. 2014.
\newblock \href {http://aclweb.org/anthology/D/D14/D14-1179.pdf} {Learning
  phrase representations using {RNN} encoder-decoder for statistical machine
  translation}.
\newblock In \emph{Proceedings of the 2014 Conference on Empirical Methods in
  Natural Language Processing, {EMNLP} 2014, October 25-29, 2014, Doha, Qatar,
  {A} meeting of SIGDAT, a Special Interest Group of the {ACL}}, pages
  1724--1734.

\bibitem[{Gollapalli et~al.(2017)Gollapalli, Li, and
  Yang}]{gollapalli2017incorporating_seqlabel}
Sujatha~Das Gollapalli, Xiaoli Li, and Peng Yang. 2017.
\newblock \href {http://aaai.org/ocs/index.php/AAAI/AAAI17/paper/view/14628}
  {Incorporating expert knowledge into keyphrase extraction}.
\newblock In \emph{Proceedings of the Thirty-First {AAAI} Conference on
  Artificial Intelligence, February 4-9, 2017, San Francisco, California,
  {USA.}}, pages 3180--3187.

\bibitem[{Gong and Zhang(2016)}]{DBLP:conf/ijcai/GongZ16}
Yuyun Gong and Qi~Zhang. 2016.
\newblock \href {http://www.ijcai.org/Abstract/16/395} {Hashtag recommendation
  using attention-based convolutional neural network}.
\newblock In \emph{Proceedings of the Twenty-Fifth International Joint
  Conference on Artificial Intelligence, {IJCAI} 2016, New York, NY, USA, 9-15
  July 2016}, pages 2782--2788.

\bibitem[{Gu et~al.(2016)Gu, Lu, Li, and Li}]{DBLP:conf/acl/GuLLL16}
Jiatao Gu, Zhengdong Lu, Hang Li, and Victor O.~K. Li. 2016.
\newblock \href {http://aclweb.org/anthology/P/P16/P16-1154.pdf} {Incorporating
  copying mechanism in sequence-to-sequence learning}.
\newblock In \emph{Proceedings of the 54th Annual Meeting of the Association
  for Computational Linguistics, {ACL} 2016, August 7-12, 2016, Berlin,
  Germany, Volume 1: Long Papers}.

\bibitem[{Hammouda et~al.(2005)Hammouda, Matute, and
  Kamel}]{hammouda2005corephrase_doc_clus_app}
Khaled~M. Hammouda, Diego~N. Matute, and Mohamed~S. Kamel. 2005.
\newblock \href {https://doi.org/10.1007/11510888\_26} {Corephrase: Keyphrase
  extraction for document clustering}.
\newblock In \emph{Machine Learning and Data Mining in Pattern Recognition, 4th
  International Conference, {MLDM} 2005, Leipzig, Germany, July 9-11, 2005,
  Proceedings}, pages 265--274.

\bibitem[{Hulth(2003)}]{Hulth2003inspec}
Anette Hulth. 2003.
\newblock \href {https://aclanthology.info/papers/W03-1028/w03-1028} {Improved
  automatic keyword extraction given more linguistic knowledge}.
\newblock In \emph{Proceedings of the Conference on Empirical Methods in
  Natural Language Processing, {EMNLP} 2003, Sapporo, Japan, July 11-12, 2003}.

\bibitem[{Hulth and Megyesi(2006)}]{hulth2006study_doc_clus_app}
Anette Hulth and Be{\'{a}}ta Megyesi. 2006.
\newblock \href {http://aclweb.org/anthology/P06-1068} {A study on
  automatically extracted keywords in text categorization}.
\newblock In \emph{{ACL} 2006, 21st International Conference on Computational
  Linguistics and 44th Annual Meeting of the Association for Computational
  Linguistics, Proceedings of the Conference, Sydney, Australia, 17-21 July
  2006}.

\bibitem[{Kim et~al.(2010)Kim, Medelyan, Kan, and Baldwin}]{Kim2010SemEval}
Su~Nam Kim, Olena Medelyan, Min{-}Yen Kan, and Timothy Baldwin. 2010.
\newblock \href {http://aclweb.org/anthology/S/S10/S10-1004.pdf} {Semeval-2010
  task 5 : Automatic keyphrase extraction from scientific articles}.
\newblock In \emph{Proceedings of the 5th International Workshop on Semantic
  Evaluation, SemEval@ACL 2010, Uppsala University, Uppsala, Sweden, July
  15-16, 2010}, pages 21--26.

\bibitem[{Kingma and Ba(2014)}]{Kingma2014Adam}
Diederik~P. Kingma and Jimmy Ba. 2014.
\newblock \href {http://arxiv.org/abs/1412.6980} {Adam: {A} method for
  stochastic optimization}.
\newblock \emph{CoRR}, abs/1412.6980.

\bibitem[{Krapivin et~al.(2009)Krapivin, Autaeu, and
  Marchese}]{Krapivin2009LargeDF}
Mikalai Krapivin, Aliaksandr Autaeu, and Maurizio Marchese. 2009.
\newblock Large dataset for keyphrases extraction.
\newblock Technical report, University of Trento.

\bibitem[{Le et~al.(2016)Le, Nguyen, and Shimazu}]{DBLP:conf/ausai/LeNS16}
Tho Thi~Ngoc Le, Minh~Le Nguyen, and Akira Shimazu. 2016.
\newblock \href {https://doi.org/10.1007/978-3-319-50127-7\_58} {Unsupervised
  keyphrase extraction: Introducing new kinds of words to keyphrases}.
\newblock In \emph{{AI} 2016: Advances in Artificial Intelligence - 29th
  Australasian Joint Conference, Hobart, TAS, Australia, December 5-8, 2016,
  Proceedings}, pages 665--671.

\bibitem[{Li et~al.(2018)Li, Tu, Yang, Lyu, and
  Zhang}]{DBLP:conf/emnlp/LiTYLZ18}
Jian Li, Zhaopeng Tu, Baosong Yang, Michael~R. Lyu, and Tong Zhang. 2018.
\newblock \href {https://aclanthology.info/papers/D18-1317/d18-1317}
  {Multi-head attention with disagreement regularization}.
\newblock In \emph{Proceedings of the 2018 Conference on Empirical Methods in
  Natural Language Processing, Brussels, Belgium, October 31 - November 4,
  2018}, pages 2897--2903.

\bibitem[{Liu et~al.(2017)Liu, Zhu, Ye, Guadarrama, and
  Murphy}]{DBLP:conf/iccv/LiuZYG017}
Siqi Liu, Zhenhai Zhu, Ning Ye, Sergio Guadarrama, and Kevin Murphy. 2017.
\newblock \href {https://doi.org/10.1109/ICCV.2017.100} {Improved image
  captioning via policy gradient optimization of spider}.
\newblock In \emph{{IEEE} International Conference on Computer Vision, {ICCV}
  2017, Venice, Italy, October 22-29, 2017}, pages 873--881.

\bibitem[{Luan et~al.(2017)Luan, Ostendorf, and
  Hajishirzi}]{luan2017scientific_seqlabel}
Yi~Luan, Mari Ostendorf, and Hannaneh Hajishirzi. 2017.
\newblock \href {https://aclanthology.info/papers/D17-1279/d17-1279}
  {Scientific information extraction with semi-supervised neural tagging}.
\newblock In \emph{Proceedings of the 2017 Conference on Empirical Methods in
  Natural Language Processing, {EMNLP} 2017, Copenhagen, Denmark, September
  9-11, 2017}, pages 2641--2651.

\bibitem[{Luong et~al.(2015)Luong, Pham, and
  Manning}]{DBLP:conf/emnlp/LuongPM15}
Thang Luong, Hieu Pham, and Christopher~D. Manning. 2015.
\newblock \href {http://aclweb.org/anthology/D/D15/D15-1166.pdf} {Effective
  approaches to attention-based neural machine translation}.
\newblock In \emph{Proceedings of the 2015 Conference on Empirical Methods in
  Natural Language Processing, {EMNLP} 2015, Lisbon, Portugal, September 17-21,
  2015}, pages 1412--1421.

\bibitem[{Medelyan et~al.(2009)Medelyan, Frank, and
  Witten}]{medelyan2009human_maui}
Olena Medelyan, Eibe Frank, and Ian~H. Witten. 2009.
\newblock \href {http://www.aclweb.org/anthology/D09-1137} {Human-competitive
  tagging using automatic keyphrase extraction}.
\newblock In \emph{Proceedings of the 2009 Conference on Empirical Methods in
  Natural Language Processing, {EMNLP} 2009, 6-7 August 2009, Singapore, {A}
  meeting of SIGDAT, a Special Interest Group of the {ACL}}, pages 1318--1327.

\bibitem[{Meng et~al.(2017)Meng, Zhao, Han, He, Brusilovsky, and
  Chi}]{DBLP:conf/acl/MengZHHBC17}
Rui Meng, Sanqiang Zhao, Shuguang Han, Daqing He, Peter Brusilovsky, and
  Yu~Chi. 2017.
\newblock \href {https://doi.org/10.18653/v1/P17-1054} {Deep keyphrase
  generation}.
\newblock In \emph{Proceedings of the 55th Annual Meeting of the Association
  for Computational Linguistics, {ACL} 2017, Vancouver, Canada, July 30 -
  August 4, Volume 1: Long Papers}, pages 582--592.

\bibitem[{Mihalcea and Tarau(2004)}]{mihalcea2004textrank}
Rada Mihalcea and Paul Tarau. 2004.
\newblock \href {http://www.aclweb.org/anthology/W04-3252} {Textrank: Bringing
  order into text}.
\newblock In \emph{Proceedings of the 2004 Conference on Empirical Methods in
  Natural Language Processing , {EMNLP} 2004, {A} meeting of SIGDAT, a Special
  Interest Group of the ACL, held in conjunction with {ACL} 2004, 25-26 July
  2004, Barcelona, Spain}, pages 404--411.

\bibitem[{Nguyen and Kan(2007{\natexlab{a}})}]{DBLP:conf/icadl/NguyenK07}
Thuy~Dung Nguyen and Min{-}Yen Kan. 2007{\natexlab{a}}.
\newblock \href {https://doi.org/10.1007/978-3-540-77094-7\_41} {Keyphrase
  extraction in scientific publications}.
\newblock In \emph{Asian Digital Libraries. Looking Back 10 Years and Forging
  New Frontiers, 10th International Conference on Asian Digital Libraries,
  {ICADL} 2007, Hanoi, Vietnam, December 10-13, 2007, Proceedings}, pages
  317--326.

\bibitem[{Nguyen and Kan(2007{\natexlab{b}})}]{Nguyen2007NUS}
Thuy~Dung Nguyen and Min{-}Yen Kan. 2007{\natexlab{b}}.
\newblock \href {https://doi.org/10.1007/978-3-540-77094-7\_41} {Keyphrase
  extraction in scientific publications}.
\newblock In \emph{Asian Digital Libraries. Looking Back 10 Years and Forging
  New Frontiers, 10th International Conference on Asian Digital Libraries,
  {ICADL} 2007, Hanoi, Vietnam, December 10-13, 2007, Proceedings}, pages
  317--326.

\bibitem[{Pasunuru and Bansal(2017)}]{DBLP:conf/emnlp/PasunuruB17}
Ramakanth Pasunuru and Mohit Bansal. 2017.
\newblock \href {https://aclanthology.info/papers/D17-1103/d17-1103}
  {Reinforced video captioning with entailment rewards}.
\newblock In \emph{Proceedings of the 2017 Conference on Empirical Methods in
  Natural Language Processing, {EMNLP} 2017, Copenhagen, Denmark, September
  9-11, 2017}, pages 979--985.

\bibitem[{Paulus et~al.(2018)Paulus, Xiong, and
  Socher}]{DBLP:conf/iclr/PaulusXS17}
Romain Paulus, Caiming Xiong, and Richard Socher. 2018.
\newblock A deep reinforced model for abstractive summarization.
\newblock In \emph{International Conference on Learning Representations
  ({ICLR})}.

\bibitem[{Porter(2006)}]{DBLP:journals/program/Porter06}
Martin~F. Porter. 2006.
\newblock \href {https://doi.org/10.1108/00330330610681286} {An algorithm for
  suffix stripping}.
\newblock \emph{Program}, 40(3):211--218.

\bibitem[{Ranzato et~al.(2015)Ranzato, Chopra, Auli, and
  Zaremba}]{DBLP:journals/corr/RanzatoCAZ15}
Marc'Aurelio Ranzato, Sumit Chopra, Michael Auli, and Wojciech Zaremba. 2015.
\newblock \href {http://arxiv.org/abs/1511.06732} {Sequence level training with
  recurrent neural networks}.
\newblock \emph{CoRR}, abs/1511.06732.

\bibitem[{Rennie et~al.(2017)Rennie, Marcheret, Mroueh, Ross, and
  Goel}]{DBLP:conf/cvpr/RennieMMRG17}
Steven~J. Rennie, Etienne Marcheret, Youssef Mroueh, Jarret Ross, and Vaibhava
  Goel. 2017.
\newblock \href {https://doi.org/10.1109/CVPR.2017.131} {Self-critical sequence
  training for image captioning}.
\newblock In \emph{2017 {IEEE} Conference on Computer Vision and Pattern
  Recognition, {CVPR} 2017, Honolulu, HI, USA, July 21-26, 2017}, pages
  1179--1195.

\bibitem[{See et~al.(2017)See, Liu, and Manning}]{DBLP:conf/acl/SeeLM17}
Abigail See, Peter~J. Liu, and Christopher~D. Manning. 2017.
\newblock \href {https://doi.org/10.18653/v1/P17-1099} {Get to the point:
  Summarization with pointer-generator networks}.
\newblock In \emph{Proceedings of the 55th Annual Meeting of the Association
  for Computational Linguistics, {ACL} 2017, Vancouver, Canada, July 30 -
  August 4, Volume 1: Long Papers}, pages 1073--1083.

\bibitem[{Sutton and Barto(1998)}]{DBLP:books/lib/SuttonB98}
Richard~S. Sutton and Andrew~G. Barto. 1998.
\newblock \href {http://www.worldcat.org/oclc/37293240} {\emph{Reinforcement
  learning - an introduction}}.
\newblock Adaptive computation and machine learning. {MIT} Press.

\bibitem[{Vaswani et~al.(2017)Vaswani, Shazeer, Parmar, Uszkoreit, Jones,
  Gomez, Kaiser, and Polosukhin}]{DBLP:conf/nips/VaswaniSPUJGKP17}
Ashish Vaswani, Noam Shazeer, Niki Parmar, Jakob Uszkoreit, Llion Jones,
  Aidan~N. Gomez, Lukasz Kaiser, and Illia Polosukhin. 2017.
\newblock \href {http://papers.nips.cc/paper/7181-attention-is-all-you-need}
  {Attention is all you need}.
\newblock In \emph{Advances in Neural Information Processing Systems 30: Annual
  Conference on Neural Information Processing Systems 2017, 4-9 December 2017,
  Long Beach, CA, {USA}}, pages 6000--6010.

\bibitem[{Wan and Xiao(2008)}]{DBLP:conf/aaai/WanX08}
Xiaojun Wan and Jianguo Xiao. 2008.
\newblock \href {http://www.aaai.org/Library/AAAI/2008/aaai08-136.php} {Single
  document keyphrase extraction using neighborhood knowledge}.
\newblock In \emph{Proceedings of the Twenty-Third {AAAI} Conference on
  Artificial Intelligence, {AAAI} 2008, Chicago, Illinois, USA, July 13-17,
  2008}, pages 855--860.

\bibitem[{Wang et~al.(2018)Wang, Yao, Tao, Zhong, Liu, and
  Du}]{DBLP:conf/ijcai/WangYTZLD18}
Li~Wang, Junlin Yao, Yunzhe Tao, Li~Zhong, Wei Liu, and Qiang Du. 2018.
\newblock \href {https://doi.org/10.24963/ijcai.2018/619} {A reinforced
  topic-aware convolutional sequence-to-sequence model for abstractive text
  summarization}.
\newblock In \emph{Proceedings of the Twenty-Seventh International Joint
  Conference on Artificial Intelligence, {IJCAI} 2018, July 13-19, 2018,
  Stockholm, Sweden.}, pages 4453--4460.

\bibitem[{Wang and Cardie(2013)}]{wang2013domain_summarize_app}
Lu~Wang and Claire Cardie. 2013.
\newblock \href {http://aclweb.org/anthology/P/P13/P13-1137.pdf}
  {Domain-independent abstract generation for focused meeting summarization}.
\newblock In \emph{Proceedings of the 51st Annual Meeting of the Association
  for Computational Linguistics, {ACL} 2013, 4-9 August 2013, Sofia, Bulgaria,
  Volume 1: Long Papers}, pages 1395--1405.

\bibitem[{Wang et~al.(2016)Wang, Zhao, and Huang}]{DBLP:conf/iconip/WangZH16}
Minmei Wang, Bo~Zhao, and Yihua Huang. 2016.
\newblock \href {https://doi.org/10.1007/978-3-319-46681-1\_15} {{PTR:}
  phrase-based topical ranking for automatic keyphrase extraction in scientific
  publications}.
\newblock In \emph{Neural Information Processing - 23rd International
  Conference, {ICONIP} 2016, Kyoto, Japan, October 16-21, 2016, Proceedings,
  Part {IV}}, pages 120--128.

\bibitem[{Wang et~al.(2019)Wang, Li, King, Lyu, and Shi}]{wang2019microblog}
Yue Wang, Jing Li, Irwin King, Michael~R Lyu, and Shuming Shi. 2019.
\newblock \href {https://arxiv.org/abs/1905.07584} {Microblog hashtag
  generation via encoding conversation contexts}.
\newblock \emph{CoRR}, abs/1905.07584.

\bibitem[{Williams(1992)}]{DBLP:journals/ml/Williams92}
Ronald~J. Williams. 1992.
\newblock \href {https://doi.org/10.1007/BF00992696} {Simple statistical
  gradient-following algorithms for connectionist reinforcement learning}.
\newblock \emph{Machine Learning}, 8:229--256.

\bibitem[{Witten et~al.(1999)Witten, Paynter, Frank, Gutwin, and
  Nevill{-}Manning}]{witten2005kea}
Ian~H. Witten, Gordon~W. Paynter, Eibe Frank, Carl Gutwin, and Craig~G.
  Nevill{-}Manning. 1999.
\newblock \href {https://doi.org/10.1145/313238.313437} {{KEA:} practical
  automatic keyphrase extraction}.
\newblock In \emph{Proceedings of the Fourth {ACM} conference on Digital
  Libraries, August 11-14, 1999, Berkeley, CA, {USA}}, pages 254--255.

\bibitem[{Wu et~al.(2016)Wu, Schuster, Chen, Le, Norouzi, Macherey, Krikun,
  Cao, Gao, and et~al.}]{DBLP:journals/corr/WuSCLNMKCGMKSJL16}
Yonghui Wu, Mike Schuster, Zhifeng Chen, Quoc~V. Le, Mohammad Norouzi, Wolfgang
  Macherey, Maxim Krikun, Yuan Cao, Qin Gao, and Klaus~Macherey et~al. 2016.
\newblock \href {http://arxiv.org/abs/1609.08144} {Google's neural machine
  translation system: Bridging the gap between human and machine translation}.
\newblock \emph{CoRR}, abs/1609.08144.

\bibitem[{Ye and Wang(2018)}]{DBLP:conf/emnlp/YeW18}
Hai Ye and Lu~Wang. 2018.
\newblock \href {https://aclanthology.info/papers/D18-1447/d18-1447}
  {Semi-supervised learning for neural keyphrase generation}.
\newblock In \emph{Proceedings of the 2018 Conference on Empirical Methods in
  Natural Language Processing, Brussels, Belgium, October 31 - November 4,
  2018}, pages 4142--4153.

\bibitem[{Yuan et~al.(2018)Yuan, Wang, Meng, Thaker, He, and
  Trischler}]{DBLP:journals/corr/diverse-keyphrase}
Xingdi Yuan, Tong Wang, Rui Meng, Khushboo Thaker, Daqing He, and Adam
  Trischler. 2018.
\newblock \href {http://arxiv.org/abs/1810.05241} {Generating diverse numbers
  of diverse keyphrases}.
\newblock \emph{CoRR}, abs/1810.05241.

\bibitem[{Zhang et~al.(2018{\natexlab{a}})Zhang, Shi, Xie, Ma, King, and
  Yeung}]{DBLP:conf/uai/ZhangSXMKY18}
Jiani Zhang, Xingjian Shi, Junyuan Xie, Hao Ma, Irwin King, and Dit{-}Yan
  Yeung. 2018{\natexlab{a}}.
\newblock \href {http://auai.org/uai2018/proceedings/papers/139.pdf} {Gaan:
  Gated attention networks for learning on large and spatiotemporal graphs}.
\newblock In \emph{Proceedings of the Thirty-Fourth Conference on Uncertainty
  in Artificial Intelligence, {UAI} 2018, Monterey, California, USA, August
  6-10, 2018}, pages 339--349.

\bibitem[{Zhang et~al.(2016)Zhang, Wang, Gong, and
  Huang}]{DBLP:conf/emnlp/ZhangWGH16}
Qi~Zhang, Yang Wang, Yeyun Gong, and Xuanjing Huang. 2016.
\newblock \href {http://aclweb.org/anthology/D/D16/D16-1080.pdf} {Keyphrase
  extraction using deep recurrent neural networks on twitter}.
\newblock In \emph{Proceedings of the 2016 Conference on Empirical Methods in
  Natural Language Processing, {EMNLP} 2016, Austin, Texas, USA, November 1-4,
  2016}, pages 836--845.

\bibitem[{Zhang et~al.(2011)Zhang, Sim, Su, and
  Tan}]{DBLP:conf/ijcai/ZhangSST11}
Wei Zhang, Yan~Chuan Sim, Jian Su, and Chew~Lim Tan. 2011.
\newblock \href {https://doi.org/10.5591/978-1-57735-516-8/IJCAI11-319} {Entity
  linking with effective acronym expansion, instance selection, and topic
  modeling}.
\newblock In \emph{{IJCAI} 2011, Proceedings of the 22nd International Joint
  Conference on Artificial Intelligence, Barcelona, Catalonia, Spain, July
  16-22, 2011}, pages 1909--1914.

\bibitem[{Zhang et~al.(2010)Zhang, Su, Tan, and
  Wang}]{DBLP:conf/coling/ZhangSTW10}
Wei Zhang, Jian Su, Chew~Lim Tan, and Wenting Wang. 2010.
\newblock \href {http://aclweb.org/anthology/C10-1145} {Entity linking
  leveraging automatically generated annotation}.
\newblock In \emph{{COLING} 2010, 23rd International Conference on
  Computational Linguistics, Proceedings of the Conference, 23-27 August 2010,
  Beijing, China}, pages 1290--1298.

\bibitem[{Zhang et~al.(2018{\natexlab{b}})Zhang, Li, Song, and
  Zhang}]{DBLP:conf/naacl/ZhangLSZ18}
Yingyi Zhang, Jing Li, Yan Song, and Chengzhi Zhang. 2018{\natexlab{b}}.
\newblock \href {https://aclanthology.info/papers/N18-1151/n18-1151} {Encoding
  conversation context for neural keyphrase extraction from microblog posts}.
\newblock In \emph{Proceedings of the 2018 Conference of the North American
  Chapter of the Association for Computational Linguistics: Human Language
  Technologies, {NAACL-HLT} 2018, New Orleans, Louisiana, USA, June 1-6, 2018,
  Volume 1 (Long Papers)}, pages 1676--1686.

\bibitem[{Zhang et~al.(2004)Zhang, Zincir{-}Heywood, and
  Milios}]{zhang2004world_summarize_app}
Yongzheng Zhang, A.~Nur Zincir{-}Heywood, and Evangelos~E. Milios. 2004.
\newblock \href
  {http://content.iospress.com/articles/web-intelligence-and-agent-systems-an-international-journal/wia00026}
  {World wide web site summarization}.
\newblock \emph{Web Intelligence and Agent Systems}, 2(1):39--53.

\end{thebibliography}
\bibliographystyle{acl_natbib}

\end{document}